\documentclass[letterpaper, 10 pt, conference]{IEEEtran}  

\usepackage[utf8]{inputenc}
\usepackage[english]{babel}
\usepackage{floatrow}
\usepackage{graphics} 
\usepackage{epsfig} 
\usepackage{mathptmx} 
\usepackage{times} 
\usepackage{cite}
\usepackage{amsmath} 
\usepackage{amssymb}  
\usepackage{graphicx}
\usepackage{mwe}
\usepackage{lipsum}%
\usepackage{booktabs}
\usepackage{multirow}
\usepackage[algoruled,boxed,lined]{algorithm2e}
\usepackage[mathscr]{euscript}
\usepackage{tabularx}
\usepackage{amsfonts}
\usepackage{url}
\usepackage{bm}
\usepackage{adjustbox}
\usepackage{footnote}
\usepackage[para,online,flushleft]{threeparttable}
\usepackage{multirow}
\newcommand{\titlename}{Traffic Accident Risk Forecasting using Contextual Vision Transformers}

\author{\IEEEauthorblockN{1\textsuperscript{st} \textbf{Khaled Saleh}}
\IEEEauthorblockA{\textit{Faculty of Engineering and IT} \\
\textit{University of Technology Sydney}\\
Sydney, Australia \\
khaled.aboufarw@uts.edu.au}
\and
\IEEEauthorblockN{2\textsuperscript{rd} \textbf{Artur Grigorev}}
\IEEEauthorblockA{\textit{Faculty of Engineering and IT} \\
\textit{University of Technology Sydney}\\
Sydney, Australia \\
ORCID: 0000-0001-6875-3568}
\and
\IEEEauthorblockN{3\textsuperscript{nd} \textbf{Adriana-Simona~Mih\u{a}i\c{t}\u{a}}}
\IEEEauthorblockA{\textit{Faculty of Engineering and IT} \\
\textit{University of Technology Sydney}\\
Sydney, Australia \\
ORCID: 0000-0001-7670-5777}}

\title{\LARGE \bf \titlename}

\begin{document}

\maketitle

\thispagestyle{empty}
\pagestyle{empty}

\begin{abstract}
Recently, the problem of traffic accident risk forecasting has been getting the attention of the intelligent transportation systems community due to its significant impact on traffic clearance. This problem is commonly tackled in the literature by using data-driven approaches that model the spatial and temporal incident impact, since they were shown to be crucial for the traffic accident risk forecasting problem. To achieve this, most approaches build different architectures to capture the spatio-temporal correlations features, making them inefficient for large traffic accident datasets. Thus, in this  work, we are proposing a novel unified framework, namely a contextual vision transformer, that can be trained in an end-to-end approach which can effectively reason about the spatial and temporal aspects of the problem while providing accurate traffic accident risk predictions. We evaluate and compare the performance of our proposed methodology against baseline approaches from the literature across two large-scale traffic accident datasets from two different geographical locations. The results have shown a significant improvement with roughly 2\% in RMSE score in comparison to previous state-of-art works (SoTA) in the literature. Moreover, our proposed approach has outperformed the SoTA technique over the two datasets while only requiring 23x fewer computational requirements.
\end{abstract}
\begin{IEEEkeywords}
traffic accident risk; risk prediction; vision transformers; deep learning
\end{IEEEkeywords}



\section{Introduction}\label{introduction}

Traffic accidents represent a major concern for cities around the world due to a significant economical and health impact to their populations. The number of vehicles has been substantially increasing during the past decades, especially in developing countries, which lead to an increase in the number of traffic accidents \cite{world2015global}. The National Highway Traffic Safety Administration (NHTSA) reports more than 5 million traffic accidents happening in the United States each year \cite{safety2013}. 
The World Health Organization also reported 1.35 million fatalities happening worldwide which resulted from traffic accidents in 2016 \cite{world2018global}.

In the past years, traffic accident research has seen an increased use of computational methods. Different problems were addressed, including: 1) traffic accident duration prediction methods, \cite{li2018overview} 2) accident detection \cite{parsa2019real}, 3) estimation of severity, and more recently, a development of spatial-temporal modelling methods have allowed to perform accident risk prediction using high-dimensional spatial, semantic and temporal data sets \cite{wang2021gsnet}. The use of such methods has enhanced the automated analysis of traffic data together with the increasing number of publicly available data sets.
Traffic accident risk prediction allows to: 1) detect high-risk areas within a traffic network, which may facilitate the decision-making inside traffic management authorities, 2) to allocate resources and assess the road design to reduce the number of accidents in the future, 3) to predict timely high-risk situations on the road and 4) to allow an implementation of risk-reducing traffic management strategies.

In the literature, the traffic accident risk forecasting problem is commonly formulated as a time-series forecasting task, where given past historical traffic accidents data for a certain city/region, along with an optional contextual information about those accidents, the objective is to forecast/predict the future traffic accident risk for that city/region. Since the nature of the traffic accident risk problem implicitly involves two types of modelling, e.g. the spatial approach (working on the affected geographic region) and the temporal approach (applied over a period of time), thus this problem is often tackled using at least two different types of model architectures.
\begin{figure}[t]
\centering
\includegraphics[keepaspectratio=true,width=\columnwidth,height=8cm]{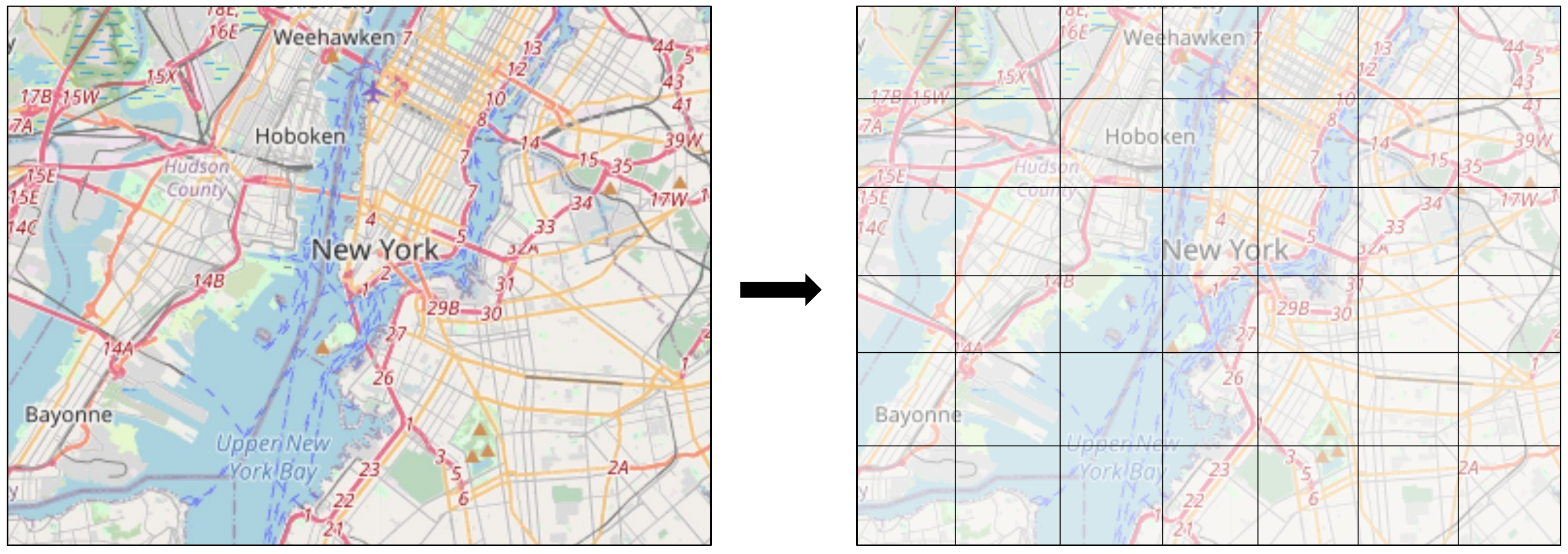}
\centering
\caption{City grid representation for our study.}
\label{fig:grid_city}
\end{figure}
\begin{figure*}[ht]
\centering
\includegraphics[keepaspectratio=true,width=\textwidth, height=9cm]{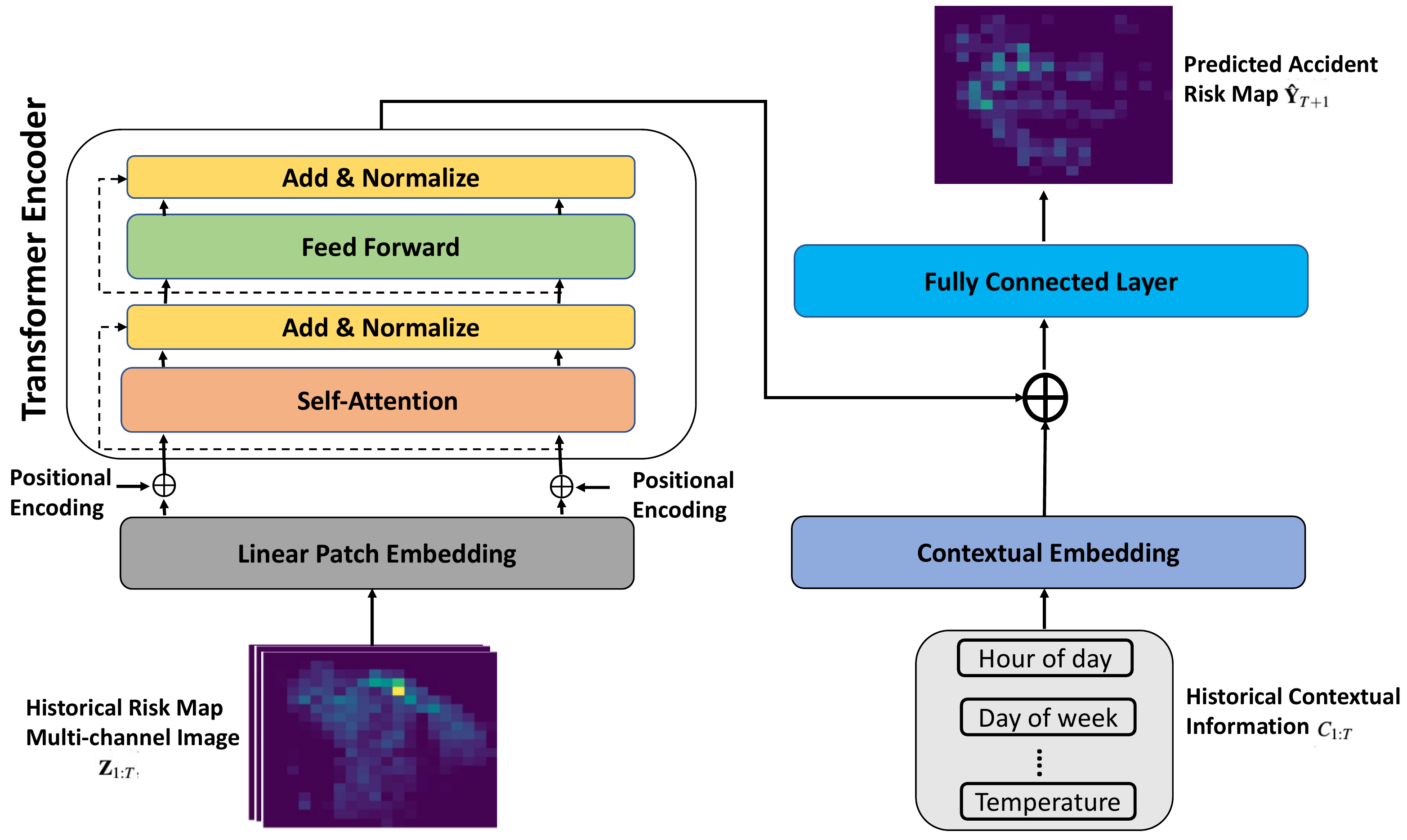}
\centering
\caption{The building blocks of our proposed C-ViT model.}
\label{fig:framework}
\end{figure*}

One of the first works on traffic accident risk prediction using Deep Learning has been performed with human mobility data using a Stack Denoise Autoencoder (SDAE) on the Japan traffic network \cite{chen2016learning}, but traffic flow and time-related matters (including periodicity) were not considered. Another research \cite{ren2017deep} relied on the LSTM network to improve the risk prediction in comparison to SAE by considering in addition the air quality, traffic flow and the weather data, represented as short-term and periodic components. \cite{zhou2020foresee} proposed also a Coarse and Fine grained prediction on the target accident risk map.
RiskOracle \cite{zhou2020riskoracle} relied on Graph-Convolution network, utilizing hierarchical coarse-to-fine modelling and proposing minute-level predictions in comparison to day-level \cite{yuan2018hetero} and hour-level \cite{chen2016learning}. In \cite{yuan2018hetero} authors have constructed over the ConvLSTM by highlighting the spatial heterogeneity problem and proposing an ensemble of region-specific ConvLSTM models (Hetero-ConvLSTM); they considered weather, the environment and the road condition in Iowa, US for over 8 years of observations, but POIs were not considered.
Semantic features, coarse and fine grained risk maps were considered in \cite{wang2021traffic}, where also Graph-convolution neural networks and attention-based LSTMs were used. A more recent work in \cite{wang2021gsnet} represents the State-of-Art (SoTA) in the field of accident risk prediction, where the authors propose a weighted loss function to address the zero-inflated issue (increase in the number of zero-risk grid cells due to the increase in the granularity of predictions) and making ensemble of models by processing semantic and geo features.
So far, risk accident prediction relied mostly upon graph-based methods and spatial-temporal modelling. While this approach worked for limited case study applications, we highly believe that in order to scale it up, this approach can benefit from using visual analysis techniques. Thus, in this work we are re-formulating the problem of traffic accident risk forecasting and we are proposing a novel approach inspired by one of the recent best performing deep learning based architectures for computer vision tasks, the vision transformers~\cite{dosovitskiy2020image}. In our proposed model we jointly model and take into account the spatio-temporal nature of the traffic accident risk forecasting problem as well as the influence of contextual information on it using a single unified end-to-end model. 

In Section~\ref{section_methodology}, a detailed description about the proposed methodology will be presented. Then, in Section~\ref{exper}, we will introduce the datasets we utilised for training and evaluating the performance of our approach, the experiments setup and the baseline approaches from the literature we compared our approach against. Finally, in Section~\ref{section_conclusion}, we conclude our paper.



\section{Methodology}\label{section_methodology}
In this section, we will first start with definitions and the problem formulation for the traffic accident risk forecasting task. Then, we will present and discuss the details of our proposed contextual vision transformer (C-ViT) model (as shown in Fig.~\ref{fig:framework}).

\subsection{Definitions}

\textbf{Grid Representation:} Given a city area bounded by certain latitude and longitude coordinates, we partition it into a grid form with $I$ rows $\times$ $J$ columns (as shown in Fig.~\ref{fig:grid_city}), where each cell share the same size.

\textbf{Traffic Accident Risk:} At any given time $t$, the traffic accident risk $Y^{i}_t$ for a grid cell $i$ is defined by the summation of the different types of traffic accidents occurred at that grid cell. Similar to~\cite{wang2021gsnet}, we have three types of traffic accidents and each one has a corresponding value, namely a minor accident has a value of 1, an injured accident a value of 2 and a fatal accident has a value of 3. For instance, if a grid cell incurred three fatal accidents and two minor accidents, the traffic accident risk for it then would be 11.

\subsection{Problem Formulation}\label{form}
In our formulation of the traffic accident prediction problem, we re-cast it as an image regression task instead of the traditional formulation as a time-series prediction task. This new formulation enables us to natively model the spatio-temporal nature of the traffic accident prediction problem in an end-to-end fashion without the need to have a combination of more than one architecture to address it. To that end, given historical observations in the form of a traffic accident risk map $\mathbf{Z}_{1:T}$, where $\mathbf{Z} \in \mathbb{R}^{I \times J}$ over time period $[1:T]$, we represent these observations as an image $X$ with a resolution of $I \times J$ and its number of channels to be $T$. Then we feed it to our proposed C-ViT model that fuse it together with the historical contextual information $C_{1:T}$ to predict/regress the future accident risk map in the next hour $\mathbf{\hat{Y}}_{T+1}$, where $\mathbf{Y} \in \mathbb{R}^{I \times J}$.

\subsection{Contextual Vision Transformer (C-ViT) Model}
Given the aforementioned formulation, we compile the traffic accident risk maps $\mathbf{Z}_{1:T}$ as a unified single image with size $T\times I \times J$, where $T$ is the number of channels, $I$ is the image's height and $J$ is the image's width, which we pass as an input to our proposed novel C-ViT model. Our C-ViT model's architecture is inspired by the recently introduced vision transformer network~\cite{dosovitskiy2020image} that has been achieving competitive results to the convolutional neural network (ConvNet) architecture for image classification tasks~\cite{wu2020visual,dosovitskiy2020image}. The main building blocks of our C-ViT model are three components, namely the historical traffic accident risk map encoding stage, the historical contextual information encoding stage and the transformer encoder stage. In the following we will analyse deeper each component.\\
\begin{figure}[ht]
\centering
\includegraphics[keepaspectratio=true,width=\columnwidth,height=8cm]{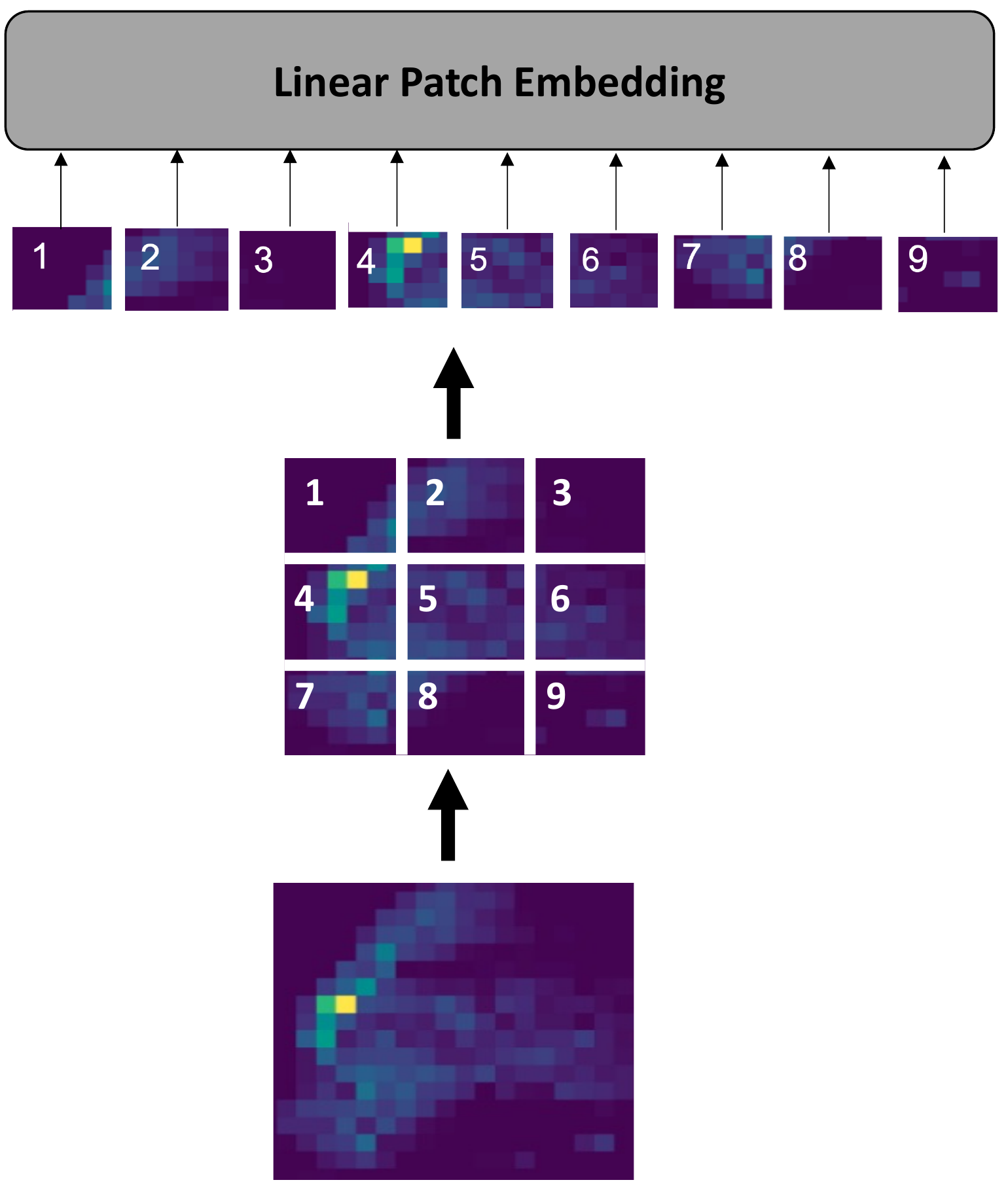}
\centering
\caption{Description of the first stage of the historical risk Map encoding. Given a unified single image $X$, it is then divided into equally-sized image patches that are passed individually to the linear patch embedding layer.}
\label{fig:image_patch}
\end{figure}

\textbf{Historical Risk Map Encoding:}
Given the historical risk maps as a unified single image $X$ with size $T \times I \times J$, we first encode it into a representation that could be easily digested and learned using our transformer encoder. As it was shown in~\cite{vaswani2017attention}, transformer encoders can work better with input data as a sequence of tokens. Thus, we divide the unified single image into a sequence of equally sub-images $X_p$ which we refer to it as an image patch sequence. We can think of the image patches as a sub-spatial regions of a number of cells within the city's grid representation that we defined in Section~\ref{form}. The rationale behind this patching process is derived by the assumption that grid cells that are spatially closer to each others will have some geographical and spatial correlations that could potentially be exploited by our model for conducting a better traffic accident risk forecasting.  

Here $X_p$ has a size of $N \times T \times P \times P$, where $P$ is the height/width of the image patch and $N$ is the total number of sequences of image patches, which is defined by $N=IJ/P^2$. The operation of dividing the unified single image into a sequence of image patches $X_p$ can be shown in Fig.~\ref{fig:image_patch}. The image patches sequence are then individually passed through a linear embedding layer which is essentially a learn-able linear projection operation in order to get a sequence of trainable flattened image patches of size $D$, which we refer to as patch embeddings. Additionally, similar to~\cite{dosovitskiy2020image}, we have an extra learnable embedding token appended before the sequence of patch embeddings to be passed to the transformer encoder and we refer to this embedding as a ``regression token''. The regression token embedding acts as an image representation which its output is transformed inside the transformer encoder into the predicted accident risk map $\mathbf{\hat{Y}}_{T+1}$.

Since the transformer encoder does not have the notion of order in its input sequence tokens, an additional position embeddings are added to each patch embedding. There are a number of pathways to define position embedding, and in our current model we follow the formulation introduced in~\cite{vaswani2017attention}. In this formulation, the position encoding $PE$ vector is defined by using a wide spectrum of frequencies of sine/cosine functions as follows:
\begin{equation}
\begin{aligned}
P E_{(a, 2 k)} &=\sin \left(a / 10000^{2 k / D}\right) \\
P E_{(a, 2 k+1)} &=\cos \left(a / 10000^{2 k / D}\right)
\end{aligned}
\end{equation}
where $a$ represents the position, and $k$ is the dimension. From the above formulation, once can conclude that for each dimension $k$ of $PE$ vector, it has a corresponding sinusoid that spans a frequency range from $2\pi$ to $10000 \cdot 2\pi$. In other words, this will allow the model to be mindful of the order in the sequential patch embedding by using unique relative positions. The dimension of the $PE$ vector is similar to the linear patch embedding layer's dimension which is $D$.\\

\textbf{Historical Contextual Information Encoding}
As discussed in Section~\ref{form}, besides the historical accident risk maps, our C-ViT model takes into account also the historical contextual information $C_{1:T}$ for the city grid representation. In our model and similar to~\cite{wang2021gsnet}, we took into account the following contextual features: 1) the time period of the day, 2) the day of the week, 3) whether the day is a holiday or not, 4) the weather condition (clear, cloud,..etc), 5) the weather temperature, and 6) traffic condition (inflow and outflow). Given those contextual features, we encode them via a learnable linear embedding layer of dimension $D$, whose output is fused together with the output from the transformer encoder via a concatenation operation.\\

\textbf{Transformer Encoder}
The main building block of our transformer encoder is the multi-head self-attention module~\cite{vaswani2017attention}. In total we have six layers inside our transformer encoder. Internally, each layer is composed of a both self-attention head and feed-forward fully connected sub-layers. Additionally, each sub-layer is followed by two residual connections and a normalisation operation. The multi-head self-attention, or the multi-scaled dot-product attention, works based on the mapping between the so-called `query' vectors and the pair (key, value) vectors. The dimension of the query and key vectors is $d_k$, where the values vector dimension is $d_v$. The attention operation itself is computed by taking the dot-product between the query and the key vectors divided by the square root of $d_k$ before finally passing them to the softmax function to get their weights by their values. Since the scaled dot-product attention operation is done multiple times, the queries, keys and values vectors are extended into matrices $Q, K, V$ respectively. The following formula is the description of how the scaled dot-product attention operation is calculated:
\begin{equation}
\mathrm{Attention}(Q, K, V) = \mathrm{softmax}(\frac{QK^T}{\sqrt{d_k}})V
\label{eq:att}
\end{equation}


\section{Experiments and Results}\label{exper}
In this section, we first present the datasets we utilised for training and evaluating the performance of our proposed approach. Then, we provide the details of the setup for our experiments, the evaluation metrics and the compared baseline approaches from the the literature. Finally, the quantitative and qualitative results of our proposed approach on real-life datasets are evaluated and discussed.

\subsection{Datasets}
In this study we use two publicly available real datasets for the traffic accident risk forecasting problem, namely NYC\footnote[1]{https://opendata.cityofnewyork.us/} and Chicago\footnote[2]{https://data.cityofchicago.org/}. As it can be seen from Table~\ref{tab:stats}, both datasets have historical traffic accidents and historical taxi trips. The historical traffic accident data contains: time, date, location (latitude and longitude), the number of causalities, the weather condition (clear, cloudy, rainy, snowy or mist), the temperature and the road segment data (i.e. road length, width and type). The NYC dataset has an additional Point of interest (POI) data regarding locations (i.e. residence, school, culture facility, recreation, social service,
transportation and commercial). The historical taxi trips include the location and times of pick-up and drop-offs and this data is used to calculate the inflow/outflow of the traffic condition in each area. 
\begin{table}[t]
\centering
\small
\begin{tabular}{ccc}
\hline Dataset & Attributes &  Range/Count \\
\hline \multirow{4}{*}{ NYC } & Reporting Duration & 1 Jan 2013 - 31 Dec 2013\\
& Accidents & 147K \\
& Taxi Trips  & 173,179K \\
& POIs & 15,625 \\
& Weathers & 8,760 \\
& Road Network & 103K \\
\hline \multirow{4}{*}{ Chicago } & Reporting Duration & 1 Feb 2016 - 30 Sep 2016\\
& Accidents & 44K \\
& Taxi Trips & 1,744K \\
& Weathers & 5,832 \\
& Road Network & 56K \\

\hline
\end{tabular}
\caption{Datasets Statistics}
\label{tab:stats}
\end{table}
\subsection{Experiment Setup}
Before we train and evaluate our proposed C-ViT model, we first pre-process the two datasets. The first pre-processing stage was to perform a grid representation by dividing each city map of the two datasets (i.e. NYC and Chicago) into equally-sized grid cells each with a dimension of ($2KM \times 2KM$). Secondly, similar to~\cite{wang2021gsnet}, we group all the accidents that happened in each grid cell based on their location over the reported duration time for each dataset (for each grid cells with no road segments/accidents, we set its traffic accident risk to zero). Thirdly, we split the data-sets into training, validation and testing. The strategy we followed for the splitting is similar to~\cite{wang2021gsnet}, where we use 60\% for training, 20\% for validation and 20\% for testing while making sure that there is no overlapping accidents based on time (i.e. no accident happened in specific grid cell on specific time is shared between the three data splits). It is worth noting that the traffic accidents periodicity according to the two datasets was set to 1 hour. Finally, each data split is standardised by a mean and standard deviation normalisation so that it could help in accelerating the training process. 

Regarding the implementation details of our C-ViT model, the size of the historical traffic risk maps $X$ was set to $7 \times 20 \times 20$ which corresponds to a total 7 historical traffic accident risks across the city grid with $I$ rows $\times$ $J$ columns of size 20. Here we chose 7 historical accident risks specifically to conform with the work done in the literature~\cite{chen2018sdcae,wang2021gsnet} for a fair comparison provided later in the paper. For each grid cell, the 7 historical accident risks comes from the most recent accident risks in past 3 hours in addition to the past accident risks in the last 4 weeks. The prediction horizon for the traffic accident risk was set to 1 (i.e next hour) similar to~\cite{chen2018sdcae,wang2021gsnet}. The hyper-parameters for our C-ViT model itself were set according to the model performance on the validation split. To that end, the $D$ dimension for the linear patch  embedding, the position embedding layer and the linear embedding layer of the historical contextual encoder was set to 64. The resolution of input patches $P$ to the patch embedding layer was set to 5. The number of self-attention heads were set to 8 and the final output fully connected layer of our C-ViT model was set to 128. 
\begin{table*} [t]
\centering
\caption{Performance evaluation of our C-ViT model against a number of baseline approaches from the literature over the NYC and Chicago datasets.}
\begin{tabular}{c|c|c|c|c|c|c}
\hline Dataset & \multicolumn{3}{|c}{ NYC } & \multicolumn{3}{|c}{ Chicago } \\
\toprule Model & RMSE $\downarrow$ & Recall $\uparrow$ & MAP $\uparrow$ & RMSE $\downarrow$ & Recall $\uparrow$ & MAP $\uparrow$ \\
\hline RNN-GRU~\cite{chung2014empirical} & 8.3375 & 28.09\% & 0.1228 & 12.6482 & 17.83\% & 0.0664 \\
SDCAE~\cite{chen2018sdcae} &  7.9774 & 30.81\% & 0.1594 & 11.3382 & 18.78\% & 0.0753\\
H-ConvLSTM~\cite{yuan2018hetero} & 7.9731 & 30.42\% &  0.1454 & 11.3033 &  18.43\% &  0.0716\\
GCN~\cite{wu2019graph} & 7.7358 & 31.78\% & 0.1623 & 11.0835 &  18.95\% & 0.0805\\
GSNet~\cite{wang2021gsnet} & 7.6151 & 33.16\% & 0.1787 & 11.3726 & 19.92\% & 0.0822\\
C-ViT (ours) & \textbf{7.0053} & \textbf{33.86\%} & \textbf{0.1875} & \textbf{9.4456} & \textbf{20.93\%} & \textbf{0.0980}\\
\bottomrule
\end{tabular}
\label{tab:results}
\end{table*}
Since we formulated the traffic accident risk prediction task as an image regression task, we have therefore optimised our C-ViT model during the training phase using a weighted mean-squared error (MSE) loss function. The reason for using the weighted MSE loss function instead of using the standard MSE loss function, is to try to combat the unbalanced nature of the traffic risk prediction problem, also known as the zero-inflated problem~\cite{bao2019spatiotemporal}. The procedure for weighting our loss function is motivated by the focal loss introduced in~\cite{lin2017focal}, where we holistically divided the total training samples into four distinctive classes based on their traffic accident risk values. Those risk values are (0, 1, 2, $\geq$3). Similar to~\cite{wang2021gsnet}, the loss function weights were set to 0.05, 0.2, 0.25 and 0.5 respectively. In total, we have trained our C-ViT model for 200 epochs using the Adam optimiser with a learning rate of 0.003 and the batch size was set to 32.
\subsection{Evaluation Metrics}
In order to evaluate the performance of our trained C-ViT model, we utilised the three commonly used metrics for the traffic accident risk prediction task~\cite{ma2018point,wang2021gsnet}, namely root mean squared error (RMSE), Recall and mean average precision (MAP). The three evaluation metrics are calculated as follows:
\begin{equation}
\mathrm{RMSE}=\sqrt{\frac{1}{N}
\sum_{n=1}^{N}\left(Y_{n}-\hat{Y}_{n}\right)^{2}}, 
\end{equation}
\begin{equation}
\text { Recall }=\frac{1}{N} \sum_{n=1}^{N} \frac{H_{n} \cap A_{n}}{\left|A_{t}\right|}, 
\end{equation}
\begin{equation}
\operatorname{MAP}=\frac{1}{N} \sum_{n=1}^{N} \frac{\sum_{j=1}^{\left|A_{t}\right|} \operatorname{PR}(j) \times \operatorname{REC}(j)}{\left|A_{n}\right|},
\end{equation}

where $N$ is the total number of samples to be evaluated, $Y_{n}, \hat{Y}_{n}$ are the ground truth and the predicted risk values for all grid cells of sample $n$ respectively. $A_n$ corresponds to the set of grid cells of sample $n$ that have an actual/true traffic accident risk values. $H_n$ corresponds to the set of grid cells within $A_n$ with the highest traffic accident risk values. On the other hand, $\operatorname{PR}(j)$ corresponds to the precision of the grid cells starting at 1 and ending at grid cell $j$. Similarly, $\operatorname{REC}(j)$ corresponds to the recall value for grid cell $j$ which is set to 1 in case there was a traffic accident risk at it and set to 0 otherwise.

Based on the definition of these three evaluation metrics, we can deduce that the lower the score of RMSE is, the better is the quality of prediction coming out of the model. On the other hand, the higher the recall and MAP scores are, the better is the accuracy of the model.

\subsection{Baselines}
We have compared the performance of our proposed C-ViT model to 5 different baseline approaches from the literature and in the following we will briefly describe each approach:
\begin{itemize}
  \item \textbf{RNN-GRU~\cite{chung2014empirical}:} This model is based on one variant of deep recurrent neural networks (RNN), the gated recurrent unit (GRU) model. This model casts the traffic accident risk forecasting problem as a time-series prediction problem and tries to model the temporal dependency among historical traffic accidents risk.
  \item \textbf{SDCAE~\cite{chen2018sdcae}:} This model is based on the stacked denoised convolution auto-encoder architecture, which focuses mainly on capturing/modelling the spatial features between different cells within a city grid area for a better prediction of the traffic accident risk.
  \item \textbf{H-ConvLSTM~\cite{yuan2018hetero}:} As the name implies, this model combines both deep convolution layers with RNN-based LSTM layers to extract the spatio-temporal features of the traffic accident risk problem by having a sliding window over the city's grid cells; this allows to have sub-regions that could potentially capture the heterogeneity among the different types of spatial regions.
  \item \textbf{GCN~\cite{wu2019graph}:} This model is a deep learning model that relies on graph convolution neural network to represent the historical traffic accident data as a graph to capture the long-term spatio-temporal dependency among historical traffic accidents risk data. 
  \item \textbf{GSNet~\cite{wang2021gsnet}:} A recent model that learns the complex spatial-temporal correlations of traffic accidents risk by using a combination of GCN, LSTM and attention mechanism. To the best of our knowledge, GSNet is currently the SOTA method on the NYC and Chicago data-sets.
\end{itemize}
\begin{table*} [t!]
\centering
\caption{Performance evaluation of our C-ViT model against a number of baseline approaches from the literature over the high frequency times of accidents in the NYC and Chicago datasets.}
\begin{tabular}{c|c|c|c|c|c|c}
\hline Dataset & \multicolumn{3}{|c}{ NYC } & \multicolumn{3}{|c}{ Chicago } \\
\toprule Model & RMSE $\downarrow$ & Recall $\uparrow$ & MAP $\uparrow$ & RMSE $\downarrow$ & Recall $\uparrow$ & MAP $\uparrow$ \\
\hline RNN-GRU~\cite{chung2014empirical} & 7.3546 & 30.76\% & 0.1301 & 9.0421 & 18.66\% & 0.0758 \\
SDCAE~\cite{chen2018sdcae} &  7.2806 & 31.22\% & 0.1536 & 8.7543 & 20.58\% & 0.1002\\
H-ConvLSTM~\cite{yuan2018hetero} & 7.2750 & 31.43\% &  0.1498 &  8.5437 & 18.93\% &  0.0770\\
GCN~\cite{wu2019graph} & 7.0958 & 33.04\% & 0.1647 & 8.4484 &  20.42\% & 0.0933\\
GSNet~\cite{wang2021gsnet} & 6.7758 & 34.15\% & 0.1769 & 8.6420 & 21.12\% & 0.1052\\
C-ViT (ours) & \textbf{6.2658} & \textbf{34.46\%} & \textbf{0.1802} & \textbf{7.0353} & \textbf{21.95\%} & \textbf{0.1247}\\
\bottomrule
\end{tabular}
\label{tab:results2}
\end{table*}
\subsection{Results}\label{results}
In Table~\ref{tab:results}, we report the results of our C-ViT model in comparison to the aforementioned baseline approaches from the literature over the total testing splits for both NYC and Chicago data-sets. As it can be noticed, our model has outperformed all the baseline approaches from the literature in terms of RMSE, recall and MAP scores over the two data-sets. It is worth noting from the results, that those models (our C-ViT, GSNet, GCN and H-ConvLSTM) which account for the spatio-temporal property of the traffic accident risk prediction problem, are the top performing approaches on the two data-sets.
\begin{figure}[ht]
\centering
\includegraphics[keepaspectratio=true,width=\columnwidth,height=8cm]{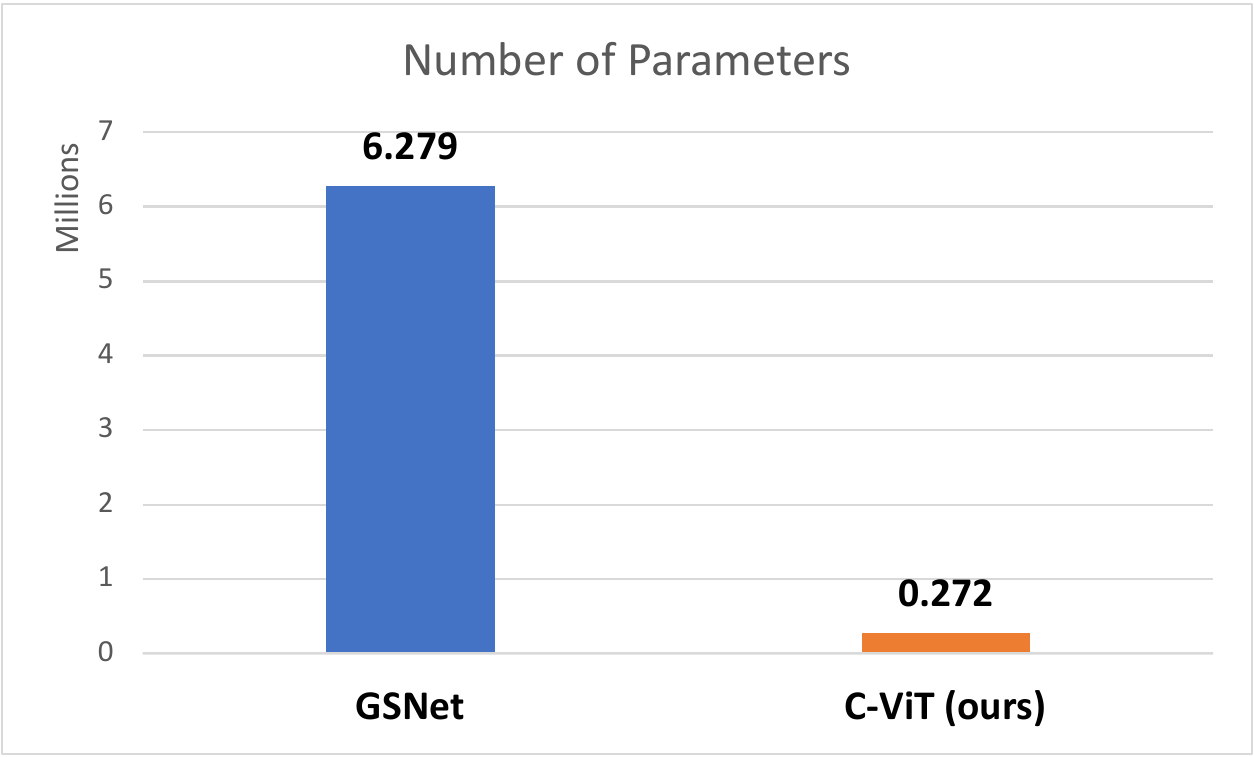}
\centering
\caption{Comparison between our proposed C-ViT model and GSNet~\cite{wang2021gsnet}, in terms of the number of training parameters.}
\label{fig:params}
\end{figure}

The closest competitor baseline approach to our C-ViT model, was the GSNet, which to the best of our knowledge, was the SOTA on the two data-sets before our proposed approach. As it can be seen, our C-ViT model has improved the RMSE, recall and MAP scores in comparison to GSNet especially across the Chicago dataset by a relatively large margin. Furthermore, our C-ViT has more competitive advantage over GSNet in terms of the efficiency. As it can be shown in Fig.~\ref{fig:params}, the number of parameters required by our C-ViT model for training are far lower than those needed for GSNet (saving more than 23x parameters) which makes our approach more suitable for real-time deployment.

In order to further evaluate the performance of our proposed C-ViT model, in Table~\ref{tab:results2} we report the RMSE, recall and MAP scores of our model when compared to all the other baseline approaches over peak hours of frequent traffic accidents that resulted from the testing split of both the NYC and Chicago data-sets. Those times of high frequency of traffic accidents are essentially during morning/evening rush hours which are within 7:00-9:00 AM and 04:00-07:00 PM. As it can be seen from the reported results, our C-ViT model continues to achieve more robust results than all other compared baseline approaches. This further prove the utility and quality of our proposed approach that it has a consistent performance across different settings.

\section{Conclusion}\label{section_conclusion}
In this work, we have presented a novel approach for the task of traffic accident risk forecasting. In our approach we re-formulated the problem as an image regression problem and introduced a unique contextual vision transformer network (C-ViT) that can efficiently model the traffic accident risk forecasting task from both spatial and temporal perspectives. The proposed approach has been evaluated on two publicly available datasets for the traffic accident risk problem. Furthermore, our proposed C-ViT model has been compared against a number of baseline approaches from the literature and it has outperformed them with a large margin while only requiring less than 23 times the number of training parameters.  
\section*{Acknowledgments}

This research is supported by the ARC LP project LP180100114. This research is funded by iMOVE CRC and supported by the Cooperative Research Centres program, an Australian Government initiative.

{\footnotesize
\bibliographystyle{IEEEtran}
\bibliography{IEEE_ITSC_2022}}


%
%
%

\end{document}